\documentclass[10pt,twocolumn,letterpaper]{article}

\usepackage[pagenumbers]{cvpr} 

\usepackage{siunitx}

\definecolor{cvprblue}{rgb}{0.21,0.49,0.74}
\usepackage[pagebackref,breaklinks,colorlinks,allcolors=cvprblue]{hyperref}
\usepackage{url}
\usepackage{newfloat}
\usepackage{listings}
\usepackage{amsmath,amsfonts,bm}
\usepackage{amssymb}
\usepackage{algorithm}
\usepackage{algpseudocode}
\usepackage{multirow}
\usepackage{enumitem}
\usepackage{listings}
\usepackage[skins,breakable]{tcolorbox}
\usepackage{cuted}     
\usepackage{mathtools} 
\usepackage{booktabs}
\usepackage{threeparttable}

\def\confName{CVPR}
\def\confYear{2026}

\usepackage{listings}
\usepackage[utf8]{inputenc}
\usepackage[export]{adjustbox}
\usepackage{lipsum}
\newcommand{\TitleLogo}{%
\raisebox{1ex}{
  \includegraphics[valign=c,width=0.08\textwidth]{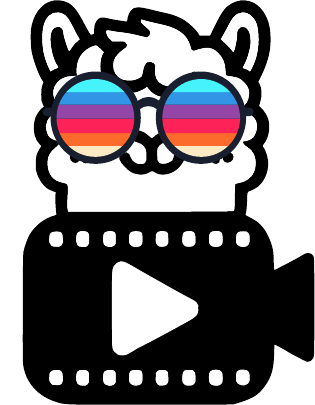}
  }
}

\newcommand{\FullTitle}{%
  \parbox[c]{0.50\textwidth}{
    {\Large Seeing the Forest and the Trees}: \\ 
    Query-Aware Tokenizer for Long-Video \\
    Multimodal Language Models
    }
}
\author{
    Siyou Li$^{1}$ \quad
    Huanan Wu$^{2}$\thanks{Corresponding author.} \quad
    Juexi Shao$^{1}$ \quad
    Yinghao Ma$^{1}$ \quad
    Yujian Gan$^{1}$ \\
    Yihao Luo$^{3}$ \quad
    Yuwei Wang$^{4}$ \quad
    Dong Nie$^{5}$ \quad
    Lu Wang$^{6}$ \quad
    Wenqing Wu$^{1,7}$ \\
    Le Zhang$^{1,8}$ \quad
    Massimo Poesio$^{1,9}$ \quad
    Juntao Yu$^{1}$ \\
    \\
    $^{1}$Queen Mary University of London, London, UK \quad
    $^{2}$University of Sheffield, Sheffield, UK \\
    $^{3}$Imperial College London, London, UK \quad
    $^{4}$Pengcheng Laboratory, Shenzhen, China\\
    $^{5}$Meta Inc, US \quad
    $^{6}$Meituan Inc, China\quad
    $^{7}$Nanjing University of Science, Nanjing, China\\
    $^{8}$University of Birmingham, Birmingham, UK \quad
    $^{9}$Utrecht University, Utrecht, Netherlands\\
    {\tt\small \{siyou.li, juntao.yu\}@qmul.ac.uk; hwu110@sheffield.ac.uk}
}

\begin{document}
\title{%
     \TitleLogo\hspace{0.2em}\FullTitle
}

\maketitle

\begin{abstract}
Despite the recent advances in the video understanding ability of multimodal large language models (MLLMs), long video understanding remains a challenge. One of the main issues is that the number of vision tokens grows linearly with video length, which causes an explosion in attention cost, memory, and latency. To solve this challenge, we present Query-aware Token Selector (\textbf{QTSplus}), a lightweight yet powerful visual token selection module that serves as an information gate between the vision encoder and LLMs. Given a text query and video tokens, QTSplus dynamically selects the most important visual evidence for the input text query by (i) scoring visual tokens via cross-attention, (ii) \emph{predicting} an instance-specific retention budget based on the complexity of the query, and (iii) \emph{selecting} Top-$n$ tokens with a differentiable straight-through estimator during training and a hard gate at inference. Furthermore, a small re-encoder preserves temporal order using absolute time information, enabling second-level localization while maintaining global coverage.

Integrated into Qwen2.5-VL, QTSplus compresses the vision stream by up to \textbf{89\%} and reduces end-to-end latency by \textbf{28\%} on long videos. The evaluation on eight long video understanding benchmarks shows near-parity accuracy overall when compared with the original Qwen models and outperforms the original model by \textbf{+20.5} and \textbf{+5.6} points respectively on TempCompass direction and order accuracies. These results show that QTSplus is an effective, general mechanism for scaling MLLMs to real-world long-video scenarios while preserving task-relevant evidence.
\end{abstract}   
\begin{figure*}[!t]
    \centering
    \includegraphics[width=0.95\textwidth]{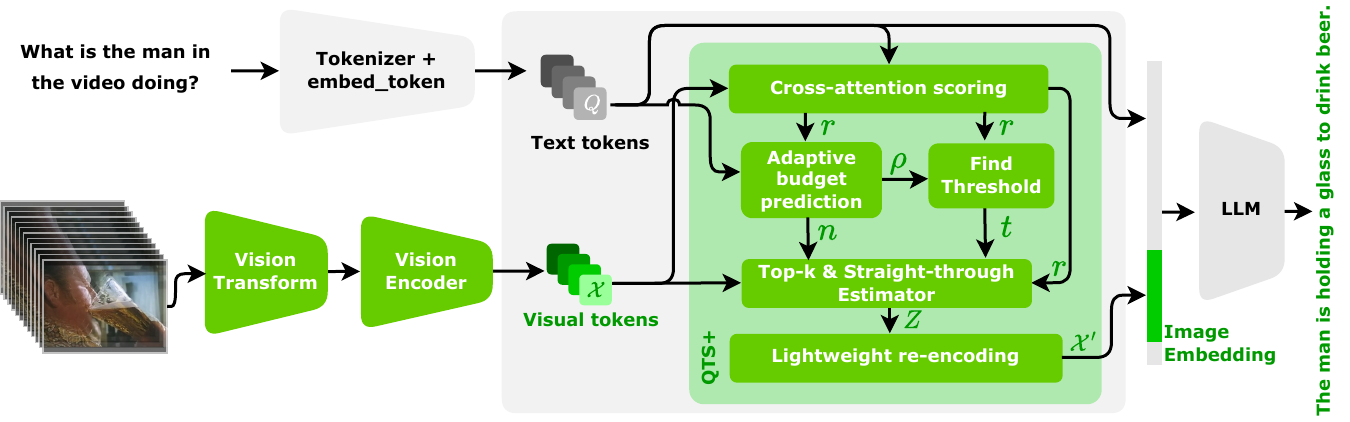}
    \caption{The illustration of our proposed QTSplus. The improvement is applied to the step of multimodal token mixture.}
    \label{example}
\end{figure*}

\section{Introduction}

Recent multimodal large language models (MLLMs) build on strong vision–language pretraining~\cite{radford2021learningtransferablevisualmodels} that extend their ability to solve image and video understanding tasks. As the ability of MLLMs grows, long video understanding becomes one of the new frontiers of multimodal AI with benchmarks like Video-MME~\cite{fu2024video}, Video-MMMU~\cite{hu2025videommmu}, LVBench~\cite{wang2024lvbench}, and MLVU~\cite{zhou2024mlvu} that emphasize tasks that require reasoning over extended temporal horizons. Understanding such long videos is crucial for applications ranging from assistive robotics and surgical coaching to copyright compliance and safety monitoring. However, most existing video–language models inherit design assumptions from short-clip recognition: they \emph{pre-truncate} the input or aggressively down-sample frames, discarding fine-grained temporal cues that matter when a question refers to “the third screw tightened” or “the moment the cake first rises”.

One native way to solve this is to input the original video without down-sampling; however, the computational and memory costs scale linearly with video duration and quickly become unrealistic when coping with multi-hour inputs that are increasingly common in user–generated content on platforms such as YouTube and Bilibili. These costs are centrally driven by the visual tokens that encode the video. Long videos generate an increasing number of these tokens, which are then used by the downstream LLM and subsequently require high computational and memory resources. As a result, \textbf{token bottlenecks} can be used as a principled solution—if they can be used to adapt to different queries.

Discrete or learned visual tokenizers and token‑pruning modules are a common method that can shrink the vision stream, but most methods apply a \emph{static} compression rate irrespective of the prompt or the temporal dispersion of evidence. This is problematic for long videos: downstream queries may hinge on a few localized moments (e.g., “\emph{When does the red light turn green?}”), while other queries require broad coverage (e.g., “\emph{Summarize the main events.}”). A fixed budget either wastes tokens on irrelevant frames or starves the model of context.

To solve this problem, we introduce a Query-aware Token Selector \textbf{QTSplus}, an intermediate module placed between the vision encoder and the language model to dynamically select the most relevant visual tokens based on the given text query. QTSplus uses cross‑attention between text and visual tokens to determine per‑token relevance; a compact controller then decides a retention fraction $\rho\!\in\![0,1]$ based on query statistics and video‑level signals (e.g. sequence length, entropy/peakedness of relevance, maximum relevance). Only visual tokens within the top fraction according to relevance are selected and fed into the language model. During training, we further use a differentiable gate with Gumbel straight‑through to enforce the pre-defined target budget. Furthermore, kept tokens are re‑encoded with absolute time information so the LLM maintains temporal consistency. This design shrinks the vision embeddings(KV-cache) and attention cost by up to 89\% without sacrificing task-relevant evidence. Together, these components allow a single MLLM to reason over multi-hour footage on commodity GPUs.

Experiments on long-video question answering (QA) and summarization benchmarks show that QTSplus-augmented Qwen2.5-VL achieves competitive or state-of-the-art results, confirming the benefits of adaptive, relevance-aware tokenisation for real-world deployments.

In summary, our main contributions are:
\begin{itemize}

\item We propose \textbf{QTSplus}, a query-aware multimodal token selector that can be plugged into existing video-language MLLMs to support efficient long-video understanding.

\item This module dynamically filters visual tokens conditioned on the textual query and absolute time encoding, substantially reducing KV-cache size while preserving task-critical evidence.

\item We construct long-video QA and single-choice question datasets via a controlled generation pipeline and evaluate QTSplus-augmented Qwen2.5-VL using a modified \texttt{lmms-eval} framework, demonstrating significant inference efficiency gains with comparable or improved task performance.
\end{itemize}

\section{Related Works}
Recent MLLMs leverage strong vision–language pre‑training and have driven progress on long‑video benchmarks. Orthogonally, efficiency research explores token pruning and routing, visual token grouping/merging, and dynamic inference. The vision encoder, often a Vision Transformer (ViT\cite{dosovitskiy2021imageworth16x16words}), typically produces a large set of patch tokens, which creates a prohibitive computational bottleneck for the LLM's self-attention mechanisms. To address this, several strategies for visual feature selection and compression have been proposed.

A dominant paradigm is \textbf{Query-Based Resampling}, which acts as an information bottleneck. This method employs a small, fixed-size set of learnable queries that distil information from the dense visual token set via cross-attention. This produces a compact, fixed-length sequence of visual embeddings that summarize the most salient features. Prominent examples of this approach include the Q-Former architecture utilized in BLIP-2\cite{li2023blip2bootstrappinglanguageimagepretraining} and the Perceiver Resampler found in Flamingo\cite{alayrac2022flamingovisuallanguagemodel}.

Alternative strategies operate directly on the visual token sequence. \textbf{Token Pruning} methods aim to reduce sequence length by identifying and discarding tokens based on saliency, redundancy, or task-relevance metrics (such as SAINT\cite{jeddi2025similarityawaretokenpruningvlm}). In contrast, \textbf{Token Merging} techniques dynamically fuse similar or spatially adjacent tokens into unified representations, progressively shortening the sequence while attempting to preserve information content. These merging mechanisms are all fundamentally designed to reduce the input sequence length for the LLM, thereby mitigating the quadratic computational cost while retaining the most critical visual information necessary for multimodal reasoning.

Our approach is motivated by the same aim to reduce the number of visual tokens, but additionally takes into account the input query and only retains visual tokens that are highly relevant to the query in question. As a result, our method can reduce the number of visual tokens more significantly than canonical token-merging methods like the ones used in ToMe\cite{bolya2023tokenmergingvitfaster} and Qwen2.5-VL\cite{bai2025qwen25vltechnicalreport}. Take the latest Qwen2.5-VL model as an example. It uses a simple MLP-based merging strategy that merges spatially adjacent tokens (e.g., n=4) into a single representation. As a result, the actual number of tokens fed into the LLM is reduced by a factor of n. Such an approach extends the model's ability from short videos to medium-length videos. However, when the input video frame rate is excessively high, the vision embedding still becomes extremely large. For example, when a 480p video input reaches 450 frames (using Qwen2.5-VL), the embedding sequence length fed to the language model reaches 136,035. This exceeds the maximum processing length of the model (131,072). In contrast, our approach can reduce the number of visual tokens by up to 89\% when compared to the vanilla Qwen2.5-VL model.
\begin{figure*}[t!]
    \centering
    \includegraphics[width=0.95\textwidth]{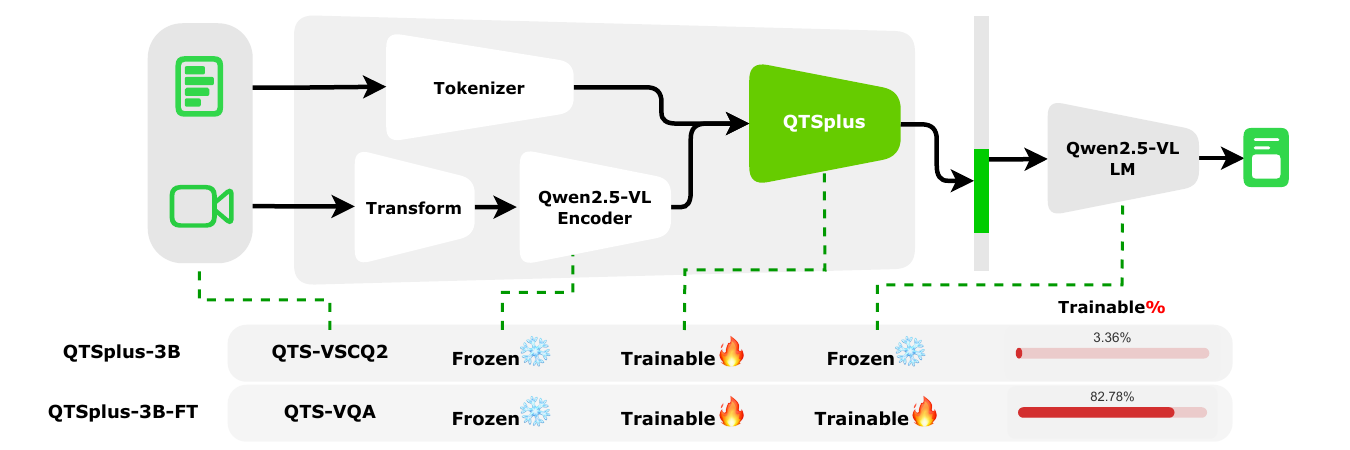}
    \caption{Comparison of the Training Processes for QTSplus-3B and QTSplus-3B-FT.}
    \label{fig:training}
\end{figure*}
\section{Method}
\subsection{Problem Set-up}

\paragraph{Inputs.}
We formalise a long video as a sequence of RGB frames
\begin{equation} 
\begin{aligned}
    \mathcal V = \bigl\{\,\mathbf f_t \in \mathbb R^{H\times W\times 3}\,\bigr\}_{t=1}^{T},
\end{aligned} 
\end{equation}
where video length  $T$ can be in the $\mathcal O(10^5)$ range (hours of sampled frames).  
At inference time, the system also receives a natural‑language prompt  
$\mathcal Q = (q_1,\dots,q_L)$ of length $L$ sub‑word tokens drawn from a text vocabulary $\mathcal V_{\text{text}}$.  
We study two tasks:

\begin{enumerate}[leftmargin=1.15em,itemsep=2pt]
    \item \textbf{Long‑Video QA}: predict an answer sequence  
        $y = (y_1,\dots,y_{|y|}) \in \mathcal V_{\text{text}}^{|y|}$  
        conditioned on the entire video and the question.
    \item \textbf{Long‑Video Summarisation}: generate a temporally coherent summary  
        $\Delta = (\delta_1,\dots,\delta_{\alpha})$ that covers salient events $\delta$ of $\mathcal V$.
\end{enumerate}

\paragraph{Vision Tokenization Pipeline.}
Each frame is first patched and projected by a frozen ViT encoder. This process yields latent features (visual tokens)
$\mathcal X = \bigl[x_1,\dots,x_M\bigr]^\top \in \mathbb R^{M\times d}$,
where the number of visual tokens is defined as $M = \tfrac{T}{\Delta t}\times\frac{HW}{P^2}$ This number can easily exceed one million tokens when the video length ($T$) is long or video resolution ($W,H$) is high. $P$ is ViT patch size, and $\Delta t$ is video sampling interval.

\paragraph{Query‑Aware Token Budget.}
To further reduce the number of selected tokens and alleviate the quadratic cost of self-attention in the downstream decoder, we introduce a \textit{Query-Aware Token Selector with Adaptive Budget}. Instead of using a fixed budget, our method dynamically determines the token budget $n \ll M$ (e.g.\ $n=256$). Given both $\mathcal{X}$ and the embedded query $q$, the selector scores and retains the top‑$n$ visual tokens most relevant to $\mathcal{Q}$. This process outputs
$\mathcal{X}' = [\,z_1,\dots,z_n]^\top \in \mathbb R^{n\times d}$.



\subsection{Query-Aware Token Selector with Adaptive Budget (QTSplus)}
\label{sec:qts}

After compressing the video stream into $M$ tokens $\mathcal{X}\!\in\!\mathbb{R}^{M\times d}$ through tokenization, the Query-Aware Token Selector (QTSplus) chooses at most $n\!\le\!n_{\max}$ query-relevant tokens for the causal decoder, thereby shrinking attention compute and the KV-cache
while preserving task evidence. QTSplus differs from prior gating mechanisms by \emph{estimate} the token budget from the query
and video statistics. During training, this budget is enforced through a differentiable, thresholded sigmoid gate, while at inference time the model switches to a hard gate Top-$n$. (See Alg.~\ref{alg:qtsplus}.)  

\begin{algorithm}[t]
\caption{Query-Aware Token Selector with Adaptive Budget (QTSplus)}
\label{alg:qtsplus}
\begin{algorithmic}[1]
\Require Visual tokens after ViT $\mathcal{X}\!\in\!\mathbb{R}^{M\times d}$ (with absolute pos.\ enc.);\,
text embeddings $\mathcal Q\!\in\!\mathbb{R}^{L\times d}$;\,
\#heads $h$ (head dim $d_h{=}d/h$);\,
temperature $\tau_s$;\,
budget head $\mathcal{B}_\psi$ with range $[\rho_{\min},\rho_{\max}]$;\,
cap $n_{\max}$;\,
mode $\in\{\texttt{train},\texttt{infer}\}$.
\Ensure Pruned visual tokens $\mathcal{X}'\!\in\!\mathbb{R}^{n\times d}$ with $n\le n_{\max}$.
\Statex
\State \textbf{// Cross-attention scoring (Sec.\,\ref{sec:qts}: Qwen2.5-style projections)}
\State Compute attention weights $\alpha\in\mathbb{R}^{h\times L\times M}$ between queries $\mathcal Q$ and keys from $\mathcal X$; \label{line:score}
\State $r \gets \max_{h,L}\alpha \in [0,1]^M$ \Comment{relevance via max over heads and text}
\Statex
\State \textbf{// Adaptive budget prediction}
\State $p \gets r / \sum_i r_i$; \, $H \gets -\sum_i p_i\log p_i$; \, $s_q \gets \tfrac{1}{L}\sum_{\ell} \mathcal Q_\ell$;\, $r_{\max}\!\gets\!\max_i r_i$
\State $\rho \gets \mathcal{B}_\psi\big(\,[s_q,\log M,r_{\max},H]\,\big)\in[\rho_{\min},\rho_{\max}]$;\quad $n\gets \min(\lceil \rho M\rceil, n_{\max})$
\Statex
\If{$\texttt{train}$} \Comment{differentiable keep/drop gate}
    \State $t \gets \textsc{FindThreshold}(r,\rho,\tau_s)$ \Comment{Alg.\,\ref{alg:findt}: solve $\sum_i \sigma((r_i-t)/\tau_s)=\rho M$}
    \State $\textit{logits} \gets \big[\,r{-}t,\,0\,\big]\in\mathbb{R}^{M\times 2}$
    \State $y \gets \textsc{GumbelSoftmax}(\textit{logits},\;\tau\!=\!\tau_s,\;\texttt{hard}\!=\!\texttt{true})$ \Comment{$y_{i,\cdot}$ is one‑hot}
    \State $s_{\text{keep}} \gets y_{:,0} \in \{0,1\}^M$ \Comment{binary keep/drop with straight‑through gradient}
    \State \textbf{if} $\sum_i s_{\text{keep},i}=0$ \textbf{then} set $s_{\text{keep},\arg\max r}\!\gets\!1$ \Comment{ensure at least one token}
    \State $Z \gets \{\,\mathcal{X}_i \mid s_{\text{keep},i}=1\,\}$ \Comment{preserve original order}
\Else \Comment{inference: hard Top-$n$}
    \State $I \gets \textsc{argsort}(r)\,[\text{take top }n\text{ ascending}]$; \quad $Z \gets \mathcal{X}[I]$
\EndIf
\Statex
\State \textbf{// optional lightweight re‑encoding}
\State $Z \gets Z + \mathrm{MHA}_h(Z)$;\quad $Z \gets Z + \mathrm{FFN}_d(Z)$ \Comment{single block with RMSNorms}
\State \Return $\mathcal{X}' \gets Z$
\end{algorithmic}
\end{algorithm}

\begin{algorithm}[t]
\caption{\textsc{FindThreshold}$(r,\rho,\tau_s)$: Newton update for solving $\sum_i \sigma((r_i - t)/\tau_s)=\rho M$}
\label{alg:findt}
\begin{algorithmic}[1]
\Require $r \in [0,1]^M$, target $\rho$, temperature $\tau_s$, iterations $J$ (e.g., $J=4\text{--}6$).
\State $t \gets \mathrm{median}(r)$
\For{$j=1$ to $J$}
    \State $s_i \gets \sigma((r_i - t)/\tau_s)$ for $i=1,\dots,M$
    \State $u \gets \sum_i s_i - \rho M$ \Comment{residual of desired sum}
    \State $g \gets -\tfrac{1}{\tau_s} \sum_i s_i (1 - s_i)$ \Comment{derivative of $u$ w.r.t.~$t$}
    \State $t \gets t - u / g$
\EndFor
\State \Return $t$
\end{algorithmic}
\end{algorithm}

\subsubsection{Cross‑Attention Scoring}
After the vision head processed the video features, we have a very long sequence of visual features. Meanwhile, the natural‑language question is embedded into a much shorter sequence of text tokens.
To decide which visual tokens matter for this particular question, we place a multi‑head cross‑attention layer \cite{vaswani2017attention} between the two streams. Let $\mathcal Q{=}[q_1,\dots,q_L]^\top \in \mathbb{R}^{L\times d}$ be the text tokens.
We compute a cross‑attention map $\alpha \in \mathbb{R}^{h\times L\times M}$ using a Qwen2.5‑compatible attention since we use Qwen2.5-VL as our base model.
The per‑token relevance is defined as
$r_i \;=\; \max_{h,\;\ell} \; \alpha_{h,\ell,i} \;\in [0,1]$ with $r \in [0,1]^M$. Intuitively, a visual token that is heavily attended by any word in the question will get a high $r$; tokens the model never looks at stay close to zero. 

\subsubsection{Adaptive Budget Estimation}
A light “budget head” $\mathcal{B}\psi$ predicts a retention fraction $\rho\in[\rho_{\min},\rho_{\max}]$ from the query and simple video statistics: $\rho=\rho_{\min}+(\rho_{\max}-\rho_{\min})\,
\sigma\!\big(w^\top\phi([\,s_q,\log M,\max_i r_i,H(p)\,])+b\big)
$. Here, $\phi(\cdot)$ is a multi‑layer MLP and $\sigma$ is the logistic function. The inputs $s_q$, $\log M$, $\max_i r_i$, and $H(p)$ are the factors used to estimate the budget, which we will discuss in detail below.

The estimated budget is then set to $n_{\text{esti}} = \big\lceil \rho, M \big\rceil $. This ratio-based approach is well-suited for variable-length inputs and ensures clean gradients during the subsequent step.The inputs for $\mathcal{B}\psi$ are chosen as follows.First, the mean query embedding, $s_q = \tfrac{1}{L}\sum_{\ell} q_\ell$, gauges semantic difficulty, allowing harder intents to request more tokens while easy lookups need fewer.Second, $\log M$ serves as a stable length cue that enforces near-monotonic budgets as the number of available visual tokens grows.Third, the peak cross-attention relevance, $\max_i r_i$, acts as a confidence spike; it implies a smaller budget is needed when salience is sharp.Fourth, the entropy of normalized relevance, $H(p) = -\sum_i p_i\log p_i$ (where $p_i=\frac{r_i}{\sum_j r_j+\varepsilon}$), is a dispersion signal that increases the budget when evidence is spread out.Together, these factors replace a fixed $n$ with a context-aware selection that preserves accuracy while reducing KV/attention cost and memory.


\paragraph{Rationale for Input Factors}

A detailed discussion of our intuition for choosing these factors is listed below:

\begin{itemize}
    \item $s_q$: \textbf{mean query embedding}. This is a compact descriptor of intent semantics—e.g., whether the prompt is a narrow lookup (“When does the door open?”) or an open-ended instruction (“Summarize the video”). The model can learn a mapping from certain semantic “modes” to typical evidence breadth. Counting or localization questions often need few, focused frames; narrative or summarization prompts need broad coverage. Using $s_q$ allows $\mathcal{B}\psi$ to raise $\rho$ for inherently wide-scope prompts and lower it for point queries.
    \item $\log M$: \textbf {log of available visual tokens after quantization}. This encourages near-monotonic budgets as the pool grows: very long videos should usually retain more absolute tokens even if the fraction $\rho$ changes slowly. Using $\log M$ (not $M$) provides scale stability so gradients do not explode with multi-hour inputs. This is a crucial consideration for the long-video regimes targeted by this paper.
    
    \item $\max_i r_i$: \textbf{peak query-vision relevance}. A sharp, high peak means the answer likely hinges on a small region or time, so a smaller budget typically suffices. Conversely, if no token stands out (low $\max_i r_i$), the evidence is diffuse, and $\rho$ should increase. This “confidence spike” heuristic thus drives smaller budgets when salience is sharp.
    \item $H(p)$: \textbf{entropy of normalized relevance}. This factor measures the dispersion of evidence. When entropy is high, relevant information is spread out, and the budget should therefore increase. Conversely, low entropy suggests concentrated information, making a smaller budget desirable.
\end{itemize}

\subsubsection{Top-n Gate}
During \textbf{training time}, we make the gate differentiable to allow gradient flow freely.
We choose the threshold $t$ so that the expected number of kept tokens matches the target budget by solving
$\sum_i \sigma\!\big((r_i-t)/\tau_s\big)=\rho M$ using a few Newton iterations initialised at the median (Alg.~\ref{alg:findt}).
Given $t$, we form per‑token binary keep/drop decisions via a \emph{Gumbel‑Softmax} \cite{jang2017categorical} \emph{straight‑through} one‑hot sample
$\text{logits}_i=\big[r_i{-}t,\;0\big], y_i=\mathrm{GumbelSoftmax}(\text{logits}_i;\tau{=}\tau_s,\;\text{hard}{=}\texttt{true}),s_{\text{keep},i}=y_{i,0}\in\{0,1\}$
with a safety fallback that forces $\max_i s_{\text{keep},i}=1$ if all zeros occur.
The kept sequence $Z$ preserves the original temporal order.

At \textbf{inference}, we use a hard Top‑$n$ over $r$ and then sort the chosen indices increasingly to keep temporal order:
$Z = \mathcal{X}[\mathrm{TopK}(r,n)]$.

\subsubsection{Lightweight Re‑encoding.}
Finally a single self‑attention re‑encoding block (RMSNorm $\to$ MHA $\to$ RMSNorm $\to$ FFN) is applied to $Z$; 
in practice we use a Qwen2.5‑style self‑attention wrapper initialised from the text model when shapes match. QTSplus returns the pruned visual tokens $Z$ together with diagnostics $(r,\rho,n)$ that we log for analysis.
(Alg.~\ref{alg:qtsplus} summarizes the steps.)

\subsubsection{Teacher Distillation}
\label{subsec:distillation}
Let $T$ denote the teacher (\emph{Qwen2.5‑VL‑3B‑Instruct}) operating on full visual embeddings, and $S$ denote our student (\emph{QTSplus‑3B} or \emph{QTSplus‑3B‑FT}) that consumes compressed embeddings produced by the \textbf{QTSplus} layer.

\paragraph{Distillation data.}
We distill from two complementary teacher‑derived corpora:
(i) the \textbf{QTS‑VSCQ2} classification set, where the correct option has been validated by $T$; and
(ii) the \textbf{QTS‑VQA} generative set, where $T$ provides free‑form answers conditioned on video and question.

\paragraph{Objectives.}
For VSCQ, $S$ predicts a distribution over options; we minimize standard cross‑entropy with the teacher‑validated correct label:
\begin{equation} 
\begin{aligned}
\mathcal{L}_{\text{mcq}} = -\log p_S(a^\star \mid q,\, g(v)),
\end{aligned} 
\end{equation}
where $q$ is the question, $v$ is the video, $a^\star$ is the correct option, and $g(\cdot)$ is the QTSplus tokenizer that converts vision features into a compact embedding stream for the language model.

For VQA, we use sequence‑level distillation: the teacher’s decoded answer $\mathbf{y}_T$ serves as the supervision target for token‑level cross‑entropy under teacher forcing:
\begin{equation} 
\begin{aligned}
\mathcal{L}_{\text{vqa}} = - \sum_{t=1}^{|\mathbf{y}_T|} \log p_S\!\big(y_{T,t}\mid \mathbf{y}_{T,<t},\, q,\, g(v)\big).
\end{aligned} 
\end{equation}
The overall objective is a simple multi‑task sum,
\begin{equation} 
\begin{aligned}
\mathcal{L} = \mathcal{L}_{\text{mcq}} + \mathcal{L}_{\text{vqa}},
\end{aligned} 
\end{equation}
applied over mixed batches from QTS‑VSCQ2 and QTS‑VQA. This formulation encourages $S$ to match the teacher’s decisions (classification) and surface forms (generation) while relying on \emph{fewer} vision tokens.

\subsection{Loss function}
We minimize a compute‑aware objective that augments supervised terms with a learned‑budget penalty. For a batch $\mathcal{B}$,

\begin{equation}
\begin{aligned}
\mathcal{L}_{\text{total}}
= \mathbb{E}_{(V,Q,y)\in\mathcal{B}}\Big[\,
  & \mathcal{L}_{\text{SFT}}(\mathcal Q,\mathcal X',y)
  + \lambda_t\,\frac{(\rho M)^2}{n_{\max}^2} \\
  & + \lambda_m\,\frac{\rho M}{n_{\max}} + \lambda_{\text{s}}(\rho-\bar\rho)^2]
\end{aligned}
\end{equation}
The two compute terms reflect quadratic attention time and linear KV memory, normalized by a practical cap $n_{\max}$. The optional quadratic prior with mean $\bar\rho$ stabilizes early training. Hyperparameters $\lambda_t,\lambda_m,\lambda_s$ trade off accuracy, efficiency. 

It is worth noting that (i) the compute penalties vanish as $\rho\!\to\!$0 and increase smoothly with $n$, giving stable gradients:
$\frac{\partial}{\partial \rho}\big[(\rho M)^2/n_{\max}^2\big]=2\rho M^2/n_{\max}^2$,
$\frac{\partial}{\partial \rho}\big[\rho M/n_{\max}\big]=M/n_{\max}$.
(ii) If a dataset‑level budget target $\bar n$ is desired, a dual variable $\alpha\ge0$ can be added to the expectation as
$\alpha(\rho M-\bar n)$ with standard ascent on $\alpha$.

\section{Experiments}
On evaluating the performance of long video understanding, we evaluate the trained model with several datasets on relevant tasks. Comparing to previous state of the art models of each dataset, including the base model we finetuned on, we can estimate the performance improve of our approach.

\subsection{Training Datasets}
\label{subsec:training_datasets}
\begin{figure*}[t]
  \centering
  \includegraphics[width=\textwidth]{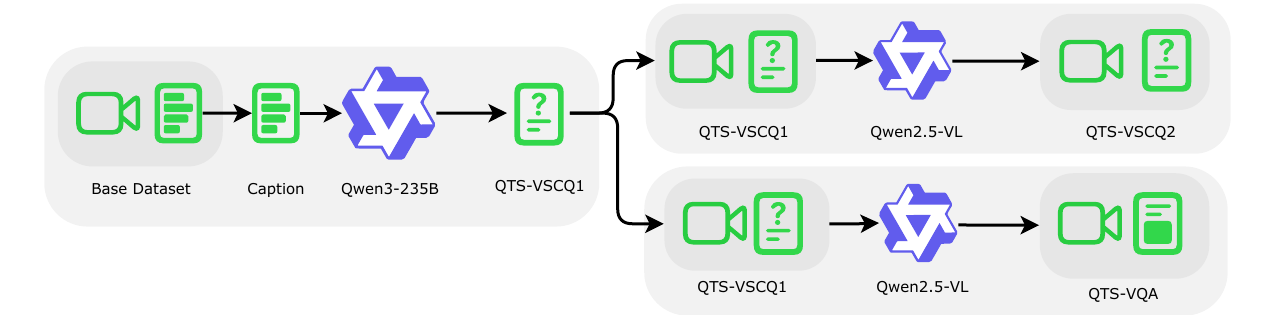}
  \caption{Data processing pipeline. VSCQ denotes visual single-choice questions. Starting from base video captions, Qwen3‑235B (text-only) synthesizes multiple-choice items (\textbf{QTS-VSCQ1}). Qwen2.5‑VL then (1) \emph{screens} QTS‑VSCQ1 by answering each item and keeping only those it judges correctly (\textbf{QTS‑VSCQ2}); and (2) \emph{generates} free-form answers given the video and question to form a visual QA corpus (\textbf{QTS‑VQA}). The two teacher‑verified sets supervise the student through distillation.}
  \label{dataset_pipeline}
\end{figure*}

\paragraph{Base corpus.}
We train on the \textbf{ShareGPTVideo} \cite{zhang2024direct} corpus containing approximately \textbf{300k} video–caption pairs. Each video is pre-segmented into frames and stored in a per-video folder to support streaming and batched loading.

\paragraph{Question synthesis \& curation.}
Fig.~\ref{dataset_pipeline} summarizes the pipeline used to transform caption supervision into training signals that are easier for a compact student to learn from. First, for every caption we prompt \textbf{Qwen3‑235B} \cite{qwen3technicalreport} to write a visual single‑choice question with options and a unique correct answer, yielding \textbf{QTS‑VSCQ1}, a VQA dataset with over \textbf{855k} multiple choices questions. Next, \textbf{Qwen2.5‑VL} acts as an automatic judge: it answers each VSCQ item from the paired video and we keep only items where the model’s prediction matches the provided answer key, forming \textbf{QTS‑VSCQ2}. Finally, we ask Qwen2.5‑VL to produce short free‑form answers (given the video and question), which we pair with the inputs to obtain the \textbf{QTS‑VQA} set for generative supervision.


\subsection{Training}
\label{subsec:training}
\paragraph{Model decomposition and wrappers.}
Following Qwen2.5‑VL’s modular design, we separate the model into a \emph{Vision Head} (vision encoder + projector to the LLM space) and a \emph{Language Model}. Each component is exported as an independent Hugging Face Transformers–compatible checkpoint and validated in isolation before integration through QTSplus.

\paragraph{QTSplus variants.}
We train two student variants(Fig.~\ref{fig:training}):
\begin{itemize}
  \item \textbf{QTSplus‑3B} (frozen Vision Head and frozen LM): Only the \emph{QTSplus} layer is trainable. This setting isolates the effect of token reduction; it shows the layer can filter and re‑encode coarse visual information well enough for downstream reasoning when supervision is framed as VSCQ.
  \item \textbf{QTSplus‑3B‑FT} (frozen Vision Head; trainable QTSplus + LM): Full fine‑tuning with teacher supervision improves the LM’s ability to \emph{read} compressed embeddings without regressing accuracy.
\end{itemize}

\paragraph{Why VSCQ for the frozen‑LM setting.}
Directly aligning a frozen LM to long free‑form captions proved brittle—style and phrasing vary across LMs, leading to plateaued loss and eventual instability. Converting captions to \emph{visual single‑choice} questions reduces linguistic variance and constrains the output space, making optimization well‑posed when only QTSplus is trainable. The teacher‑screening step ensures that, given sufficient (uncompressed) evidence, the base model can solve the item; the student then learns to preserve just enough evidence through compressed embeddings to reach the same answer.

\paragraph{QTSplus configuration.}
Unless otherwise stated, we set the cross‑attention \emph{scoring} depth to \textbf{1} and the \emph{re‑encoding} depth to \textbf{2}. To accelerate training, we reuse the multi‑head attention configuration and weights from \textbf{Qwen2.5‑VL-3B‑Instruct} inside the re‑encoder and enable FlashAttention for both training and inference.

\paragraph{Parameter counts and compute.}
\textbf{QTSplus‑3B‑FT} contains \textbf{3{,}215{,}495{,}937} trainable parameters (\textbf{82.78\%} of total model weights). \textbf{QTSplus‑3B} trains \textbf{129{,}557{,}249} parameters (\textbf{3.36\%}). Experiments were run on a server with eight NVIDIA RTX~5090 GPUs, 720\,GB RAM, and a 200‑core Intel Xeon processor.

\subsection{Evaluations on Overheads}
\label{subsec:evaluations}
\begin{figure*}[t]
  \centering
  \includegraphics[width=\textwidth]{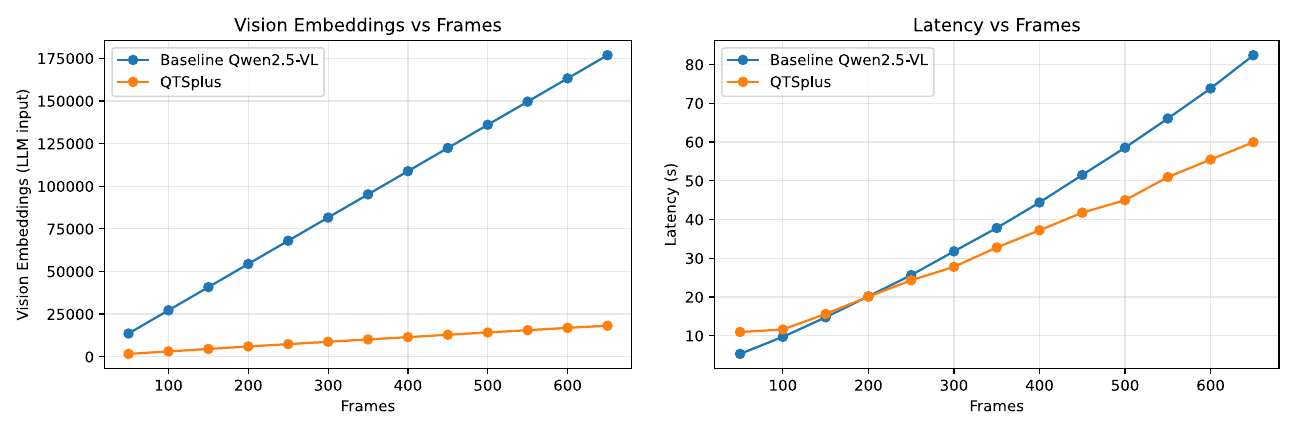}
  \caption{Scaling characteristics on a single A100 GPU (FlashAttention\cite{dao2022flashattentionfastmemoryefficientexact} enabled). \textbf{Left:} number of vision embeddings forwarded to the LM versus frame count. \textbf{Right:} end‑to‑end latency to first complete output versus frame count. QTSplus‑3B‑FT consistently reduces the embedding budget and latency relative to Qwen2.5‑VL‑3B‑Instruct.}
  \label{fig:system_load}
\end{figure*}

QTSplus reduces the vision stream almost an order of magnitude while preserving answer quality. At $\sim$600 frames, the number of vision embeddings drops from $\approx$180k to $\approx$20k (a \textbf{$\approx$89\%} reduction), and wall‑clock latency decreases from $\approx$83\,s to $\approx$60\,s (a \textbf{$\approx$28\%} reduction) on a single A100 (Fig.~\ref{fig:system_load}). The savings scale linearly with video length because the learned retention fraction $\rho$ adapts to query difficulty and evidence dispersion rather than duration alone.

\paragraph{Setup.}
We compare \textbf{QTSplus‑3B‑FT} to \textbf{Qwen2.5‑VL‑3B‑Instruct} under identical decoding settings on a \emph{single} NVIDIA A100. We vary the number of video frames while measuring (i) the count of visual embeddings actually consumed by the LM and (ii) end‑to‑end inference latency.

\paragraph{Results.}
Across frame counts, QTSplus reduces the visual embedding input to the LM by \textbf{89.7\%}, and shortens wall‑clock inference by \textbf{27.2\%}. The left panel of Fig.~\ref{fig:system_load} visualizes the near‑linear scaling of embeddings with frames, with QTSplus showing a markedly lower slope; the right panel shows corresponding latency reductions under the same conditions. Fig.~\ref{fig:system_load} provides the full curves. 


\begin{table}[ht]
  \centering
  \begin{threeparttable}
  \caption{Performance on Tasks}
  \label{tab:results}
  \renewcommand\arraystretch{1.15}
  \setlength{\tabcolsep}{8pt}
      \begin{tabular}{lp{0.12\linewidth}p{0.12\linewidth}p{0.12\linewidth}}
        \toprule
        \textbf{Datasets} &
        \textbf{Qwen3B} &
        \textbf{QTS3B} &
        \textbf{QTS3Bf} \\
        \cmidrule(lr){1-4}
        \multicolumn{4}{c}{\textit{General Multimodal Tasks}} \\
        \cmidrule(lr){1-4}
        TC$_{avg\_accuracy}$ & \underline{66.27} & 64.75 & \textbf{67.09} \\
        TC$_{action\_accuracy}$ & \textbf{97.93} & 97.04 & \underline{97.63} \\
        TC$_{attribute\_change\_acc}$ & \textbf{74.31} & 69.20 & \underline{73.61} \\
        TC$_{direction\_accuracy}$ & 43.58 & \textbf{64.07} & \underline{43.88} \\
        TC$_{order\_accuracy}$ & 63.91 & \textbf{69.54} & \underline{68.21} \\
        TC$_{speed\_accuracy}$ & \underline{51.42} & 49.27 & \textbf{52.05} \\
        \cmidrule(lr){1-4}
        \multicolumn{4}{c}{\textit{Video Understanding Tasks}} \\
        \cmidrule(lr){1-4}
        Video-MME & \textbf{57.81} & \underline{57.07} & 55.78 \\
        LVBench & \textbf{39.44} & 37.90 & \underline{38.48} \\
        MLVU$_{test}$ & \textbf{31.78} & 27.37 & \underline{30.75} \\
        VM$_{adaptation}$ & 28.33 & \underline{30.33} & \textbf{33.00} \\
        VM$_{perception}$ & \textbf{56.33} & \underline{55.67} & 53.00 \\
        VM$_{comprehension}$ & \textbf{41.00} & 37.00 & \underline{38.00} \\
        MB$_{action\_count}$ & \underline{58.50} & \textbf{62.50} & 55.50 \\
        MB$_{action\_sequence}$ & \textbf{78.00} & 69.00 & \underline{70.50} \\
        MB$_{unexpected\_action}$ & \underline{79.50} & \textbf{82.50} & \underline{79.50} \\
        MB$_{fine\_grained\_action}$ & \underline{46.00} & 48.00 & \textbf{50.00} \\
        MB$_{object\_existence}$ & \textbf{93.50} & 92.00 & \underline{93.00} \\
        MB$_{object\_interaction}$ & \textbf{74.50} & 72.00 & \underline{73.50} \\
        MB$_{scene\_transition}$ & \textbf{91.50} & \underline{91.00} & \underline{91.00} \\
        MB$_{moving\_attribute}$ & \textbf{96.50} & 91.50 & \underline{92.50} \\
        \bottomrule
      \end{tabular}
    \begin{tablenotes}[flushleft,para] 
      \scriptsize
      \item[]
      \begin{tabular*}{\linewidth}{@{\extracolsep{\fill}}ll ll}
        \textbf{Model abbreviations:} & \textbf{Task abbreviations:} \\
        Qwen3B = Qwen2.5-VL-3B-Instruct & TC = TempCompass Multiple Choices \\
        QTS3B = QTSplus-3B              & VM = Video-MMMU \\
        QTS3Bf = QTSplus-3B-FT          & MB = MVBench \\
      \end{tabular*}
    \end{tablenotes}

  \end{threeparttable}
\end{table}

\subsection{Evaluations on Open Datasets}
\label{subsec:evaluations_on_datasets}

We apply the tokenizer to Qwen2.5-VL 3B and 7B to verify performance improvements.

\subsubsection{Framework}

Since Qwen2.5-VL has been extensively validated on public datasets, we mainly evaluate on a subset of accessible datasets. However, with updates to dependencies and evaluation tools such as transformers, prior model baselines have shifted. Therefore, we compare the original Qwen2.5-VL 3B/7B with their counterparts using the QTSplus tokenizer on the same evaluation tasks.

Our goal is: after adopting the QTSplus tokenizer, model performance should not degrade (and preferably improve), while tokenization efficiency is optimized so the model can process longer video inputs under a smaller context budget. Accordingly, we compare model performance under existing evaluation criteria.

We use the popular open-source multimodal evaluation framework lmms-eval\cite{zhang2024lmmsevalrealitycheckevaluation}. This framework aggregates many popular multimodal benchmarks and provides most datasets, enabling efficient multi-dataset, multi-model testing. The following results are based on datasets and subsets integrated by the framework. To support inference for multimodal LMs using QTSplus, we made minor modifications to the framework, including some bug fixes and adaptation for QTSplus models.

Some datasets stored in the lmms-eval repository may differ from their official versions; for compatibility, certain implementation details may also vary. We observed feedback that results obtained with this tool can diverge from leaderboard numbers. We believe that the underlying LLM inference environment evolves over time, making it hard to exactly match results from a year ago. For example, some test suites (e.g., TempoCompass\cite{chen2025mega-bench}) use GPT-3.5 Turbo for assisted verification. Since this service is inaccessible in our region, we use the latest QWen3-235B-A22B-Instruct as a substitute. As these LLMs are only used to assess answer alignment, the impact on results should be limited.

Nonetheless, we evaluate baseline and improved models under the same environment and criteria, and compare using the latest results, which suffices to validate the improvements. In addition, our evaluation code will be opensourced and reproducible, ensuring stability and credibility.

\subsection{Evaluations on Open Datasets}
\label{subsec:evaluations_on_datasets}

We evaluate QTSplus with Qwen2.5‑VL (3B) under the same protocol as the teacher and report accuracy on a representative subset of public benchmarks using the open‑source \texttt{lmms‑eval} framework (with minor fixes for long‑video inputs). Where repository snapshots differ slightly from the original leaderboards, we evaluate both baseline and QTSplus under the \emph{same} code and data for a fair comparison.

\paragraph{Benchmarks.}
\textbf{TempoCompass}~\cite{chen2025mega-bench} checks the fine grained understanding of videos. 
\textbf{Video‑MME}~\cite{fu2024video} measures broad video understanding across scenarios.  
\textbf{Video‑MMMU}~\cite{hu2025videommmu} targets diverse, expert‑level tasks and reports adaptation/perception/comprehension splits (we follow the adaptation metric).  
\textbf{LVBench}~\cite{wang2024lvbench} emphasizes long videos up to two hours.  
\textbf{MLVU}~\cite{zhou2024mlvu} focuses on multimodal long‑video understanding with global and local reasoning.  
\textbf{MVBench}~\cite{li2024mvbenchcomprehensivemultimodalvideo} contains 20 temporal sub‑tasks (e.g., action count/sequence, object interaction, scene transition).  

\subsection{Generalization}

\begin{table}[h]
\centering
\resizebox{\columnwidth}{!}{%
\begin{tabular}{lcccc}
\toprule
\textbf{Model} & \textbf{VideoMME} & \textbf{MLVU} & \textbf{LongVB} & \textbf{LVBench} \\
\midrule
LLaVA-Video-7B & 54.11 & 63.78 & 49.47 & 40.03 \\
+ QTSplus & 52.89 & 60.93 & 13.51* & 36.67 \\
\midrule
InternVL2.5-8B & 54.78 & 64.93 & 48.53 & 40.87 \\
+ QTSplus & \textbf{54.22} & \textbf{60.70} & \textbf{39.75} & \textbf{36.99} \\
\bottomrule
\end{tabular}%
}
\caption{Generalization of QTSplus to LLaVA and InternVL. \\(*) indicates lack of specific format tuning during rebuttal.}
\label{tab:generalization}
\vspace{-1.5em}
\end{table}

To demonstrate that QTSplus is model-agnostic, we extended it to two additional state-of-the-art video-LLMs: \textbf{LLaVA-Video-7B} and \textbf{InternVL2.5-8B}. We applied QTSplus (frozen vision encoder and language model, trainable adapter) without extensive tuning. 

As shown in \Cref{tab:generalization}, QTSplus successfully generalizes. On InternVL2.5, QTSplus retains $\approx$99\% of performance on VideoMME and LVBench while reducing the visual token count by $\approx$89\%. The drop on LongVB for LLaVA is due to the lack of instruction tuning on that specific format in the limited rebuttal time, yet the high retention on VideoMME proves the module effectively selects informative tokens across different backbones.

\begin{table}[ht]
  \centering
  \begin{threeparttable}
  \caption{Performance of Different Components}
  \label{tab:ablation}
  \renewcommand\arraystretch{1.15}
  \setlength{\tabcolsep}{6pt}
    \begin{tabular}{lp{0.12\linewidth}p{0.12\linewidth}p{0.12\linewidth}}
      \toprule
      \textbf{Datasets} &
      \textbf{UNIF} &
      \textbf{nREENC} &
      \textbf{QTS3B} \\
      \cmidrule(lr){1-4}
      \multicolumn{4}{c}{\textit{General Multimodal Tasks}} \\
      \cmidrule(lr){1-4}
      TC$_{action\_accuracy}$ & 94.67 & \underline{95.56} & \textbf{96.75} \\
      TC$_{attribute\_change\_acc}$ & 57.99 & \underline{60.76} & \textbf{67.01} \\
      TC$_{avg\_accuracy}$ & 59.94 & \underline{61.08} & \textbf{64.18} \\
      TC$_{direction\_accuracy}$ & \underline{38.21} & \textbf{39.10} & \textbf{39.10} \\
      TC$_{order\_accuracy}$ & 58.61 & \underline{61.26} & \textbf{67.88} \\
      TC$_{speed\_accuracy}$ & \underline{48.90} & 47.63 & \textbf{49.84} \\
      \cmidrule(lr){1-4}
      \multicolumn{4}{c}{\textit{Video Understanding Tasks}} \\
      \cmidrule(lr){1-4}
      Video-MME & 47.63 & \underline{51.48} & \textbf{53.56} \\
      LVBench & 32.92 & \underline{33.38} & \textbf{36.86} \\
      MLVU$_{test}$ & \underline{13.45} & 9.82 & \textbf{22.51} \\
      VM$_{adaptation}$ & \underline{25.33} & \underline{25.33} & \textbf{25.67} \\
      VM$_{comprehension}$ & \underline{34.67} & 33.67 & \textbf{35.67} \\
      VM$_{perception}$ & 39.33 & \underline{40.67} & \textbf{47.00} \\
      MB$_{action\_antonym}$ & 79.00 & \underline{79.50} & \textbf{80.00} \\
      MB$_{action\_count}$ & \underline{53.50} & \underline{53.50} & \textbf{62.00} \\
      MB$_{action\_localization}$ & 41.50 & \textbf{44.00} & \underline{42.00} \\
      MB$_{action\_prediction}$ & 57.50 & \underline{59.50} & \textbf{63.50} \\
      MB$_{action\_sequence}$ & \underline{66.50} & 65.00 & \textbf{67.50} \\
      MB$_{character\_order}$ & 56.00 & \underline{68.50} & \textbf{71.50} \\
      MB$_{counterfactual\_inference}$ & 56.50 & \underline{67.00} & \textbf{74.00} \\
      MB$_{egocentric\_navigation}$ & 39.00 & \underline{39.50} & \textbf{41.00} \\
      MB$_{episodic\_reasoning}$ & 49.50 & \underline{51.50} & \textbf{56.50} \\
      MB$_{fine\_grained\_action}$ & \textbf{47.00} & \textbf{47.00} & \underline{46.50} \\
      MB$_{fine\_grained\_pose}$ & 50.50 & \underline{54.50} & \textbf{60.00} \\
      MB$_{moving\_attribute}$ & 75.00 & \underline{85.50} & \textbf{89.50} \\
      MB$_{moving\_count}$ & 56.00 & \underline{60.00} & \textbf{65.00} \\
      MB$_{moving\_direction}$ & 37.00 & \underline{40.50} & \textbf{41.50} \\
      MB$_{object\_existence}$ & 74.50 & \underline{85.00} & \textbf{86.00} \\
      MB$_{object\_interaction}$ & \underline{66.00} & \textbf{68.00} & \underline{67.50} \\
      MB$_{object\_shuffle}$ & 32.50 & \underline{33.50} & \textbf{34.00} \\
      MB$_{scene\_transition}$ & \underline{89.00} & 88.00 & \textbf{89.50} \\
      MB$_{state\_change}$ & 53.00 & \underline{53.50} & \textbf{55.00} \\
      MB$_{unexpected\_action}$ & 76.00 & \underline{81.00} & \textbf{82.00} \\
      \bottomrule
    \end{tabular}
    \begin{tablenotes}[flushleft,para] 
      \scriptsize
      \parbox{\linewidth}{
        \textbf{Model abbreviations:}
          UNIF = Replaced the whole QTSplus layer with uniform sampling while keeping other components unchanged; 
          nREENC = QTSplus-3B's QTSplus layer, disabling the reencode layer, keep other components;
          QTS3B = QTSplus-3B with default configuration.
        \textbf{The maximum number of vision embeddings in the above model is set to 256.}
        \textbf{Task abbreviations:}
          TC = TempCompass Multiple Choices; 
          VM = Video-MMMU;
          MB = MVBench.
      }
    \end{tablenotes}
  \end{threeparttable}
\end{table}

\section{Ablation Study}
\label{app:ablation}

\paragraph{Protocol and variants}
We ablate the QTSplus components using a stricter fixed cap of $\texttt{max}=256$ retained visual embeddings per instance to demonstrate our selector's effectiveness in an extreme scenario. In contrast, both QTSplus-3B and QTSplus-3B-FT (Table~\ref{tab:results}) utilize a cap of $\texttt{max}=25,600$ to balance efficiency and effectiveness.
\begin{enumerate}
  \item \textbf{UNIF}: replace the whole QTSplus layer with uniform sampling.
  \item \textbf{nREENC}: QTSplus without the lightweight re-encoder (token-select only).
  \item \textbf{QTS3B}: full configuration (query-aware selector \(+\) re-encoder).
\end{enumerate}
Table~\ref{tab:ablation} in the appendix reports detailed results for general tasks and MVBench subtasks under this setting.

\paragraph{Aggregate effects}
Relative to \textbf{UNIF}, the full model (\textbf{QTS3B}) improves the six general TempoCompass metrics by an average of \textbf{+4.41} points and the six aggregate video benchmarks (\emph{Video-MME}, \emph{LVBench}, \emph{MLVU\textsubscript{test}}, \emph{Video-MMMU} adaptation/perception/comprehension) by \textbf{+4.66} points. Against \textbf{nREENC}, the gains are \textbf{+3.23} (general) and \textbf{+4.49} (aggregate video). Representative deltas (UNIF \(\to\) QTS3B; nREENC \(\to\) QTS3B) include:
\begin{itemize}
  \item \emph{Video-MME}: \textbf{+5.93}; \textbf{+2.08}.
  \item \emph{LVBench}: \textbf{+3.94}; \textbf{+3.48}.
  \item \emph{MLVU\textsubscript{test}}: \textbf{+9.06}; \textbf{+12.69}.
  \item TempCompass \textbf{order}: \textbf{+9.27}; \textbf{+6.62}, and \textbf{attribute-change}: \textbf{+9.02}; \textbf{+6.25}.
\end{itemize}

\paragraph{Fine-grained MVBench analysis.}
The re-encoder consistently strengthens tasks that require strict temporal alignment or spatially coherent aggregation of sparse cues. Salient examples (UNIF \(\to\) QTS3B):
\begin{itemize}
  \item \textbf{Character order} \(\mathbf{+15.5}\) (56.0 \(\to\) 71.5) and \textbf{counterfactual inference} \(\mathbf{+17.5}\) (56.5 \(\to\) 74.0), indicating improved modeling of long-range dependencies and the ability to separate competing visual evidence.
  \item \textbf{Fine-grained pose} \(\mathbf{+9.5}\) (50.5 \(\to\) 60.0) and \textbf{moving attribute} \(\mathbf{+14.5}\) (75.0 \(\to\) 89.5), showing benefits on subtle, localized visual attributes once tokens are re-encoded with absolute time.
  \item \textbf{Action count} \(\mathbf{+8.5}\) and \textbf{episodic reasoning} \(\mathbf{+7.0}\), reflecting better temporal consolidation across scattered snippets.
\end{itemize}
Regressions are rare and small. The only consistent dip is \textbf{fine-grained action} (\(-0.5\) vs.\ both UNIF and nREENC), suggesting that extremely local appearance cues can occasionally be over-compressed; increasing the per-instance budget or adding a per-category minimum may mitigate this. Compared with \textbf{nREENC}, the re-encoder particularly helps where order matters (e.g., \textit{character order} \(+3.0\)) and where pose/geometry must be integrated across frames (e.g., \textit{fine-grained pose} \(+5.5\)).

\paragraph{Summary}
(i) Query-aware selection is strictly better than uniform retention across nearly all metrics. (ii) The lightweight re-encoder yields the majority of the remaining gains by restoring temporal coherence after selection, especially for order, pose, and counterfactual reasoning. (iii) The ablation confirms that QTSplus’ two-stage design—query-conditioned gating followed by temporal re-encoding—is necessary to recover or exceed teacher-level accuracy under a strict token budget of \(256\) while enabling the efficiency gains reported above.

\paragraph{Main results.}
We summarize results in Tab.~\ref{tab:results}, QTSplus‑3B achieves near‑parity overall with the teacher while being far more efficient: on Video‑MME it reaches \textbf{57.07} vs \textbf{57.81}; on LVBench \textbf{37.90} vs \textbf{39.44}; and on MLVU‑Test \textbf{27.37} vs \textbf{31.78}. Notably, it \emph{improves} TempCompass direction by \textbf{+20.5} points and order by \textbf{+5.6} points, and increases Video‑MMMU adaptation by \textbf{+2.0}. Averaged over the general‑tasks block, QTSplus‑3B is \textbf{+2.74} points; over the video‑tasks block it is $-1.35$ points relative to the teacher (still within a narrow band).

The QTSplus‑3B‑FT variant, which fine‑tunes the LLM to read compressed embeddings, further improves Video‑MMMU adaptation to \textbf{33.0} $+4.7$ over teacher) and MB fine‑grained action to \textbf{50.0} $+4.0$). 

Drops on some dense‑evidence tasks (e.g., Video‑MME and MLVU) remain modest ($\leq 3$ points), suggesting that additional budget or curriculum on global‑coverage questions could close the gap.

\paragraph{Takeaways.}
Query‑conditioned token selection preserves task‑critical snippets and helps on temporally focused queries (e.g., direction/order in TempCompass, MVBench unexpected/fine‑grained action), while keeping overall accuracy comparable to the teacher at a fraction of the compute and memory.
\section{Conclusion}

We introduced \textbf{QTSplus}, a query‑aware tokenization module for long‑video MLLMs. By scoring vision tokens with cross‑attention and \emph{estimating} a per‑instance retention budget, QTSplus keeps only the tokens that matter for the current question while preserving temporal structure through lightweight re‑encoding. Integrated into Qwen2.5‑VL, QTSplus compresses the vision stream by \textbf{$\approx 89\%$} and reduces latency by \textbf{$\approx 28\%$} on long videos, yet matches or improves accuracy on several public benchmarks, including large gains on temporally focused metrics.

Our results indicate that adaptive, relevance‑aware tokenization is a practical path to scaling MLLMs to hour‑level inputs under realistic compute and memory limits. Future work includes (i) curriculum and budget scheduling for tasks requiring broader coverage, (ii) streaming and continual inference where the budget evolves over time, and (iii) extending QTSplus to multi‑query interactions and multi‑camera inputs. We will release code and data to encourage reproducible research on efficient long‑video understanding.


\clearpage
\setcounter{page}{1}
\maketitlesupplementary


\appendix
\begin{figure*}[!ht]
  \centering
  \includegraphics[width=\linewidth]{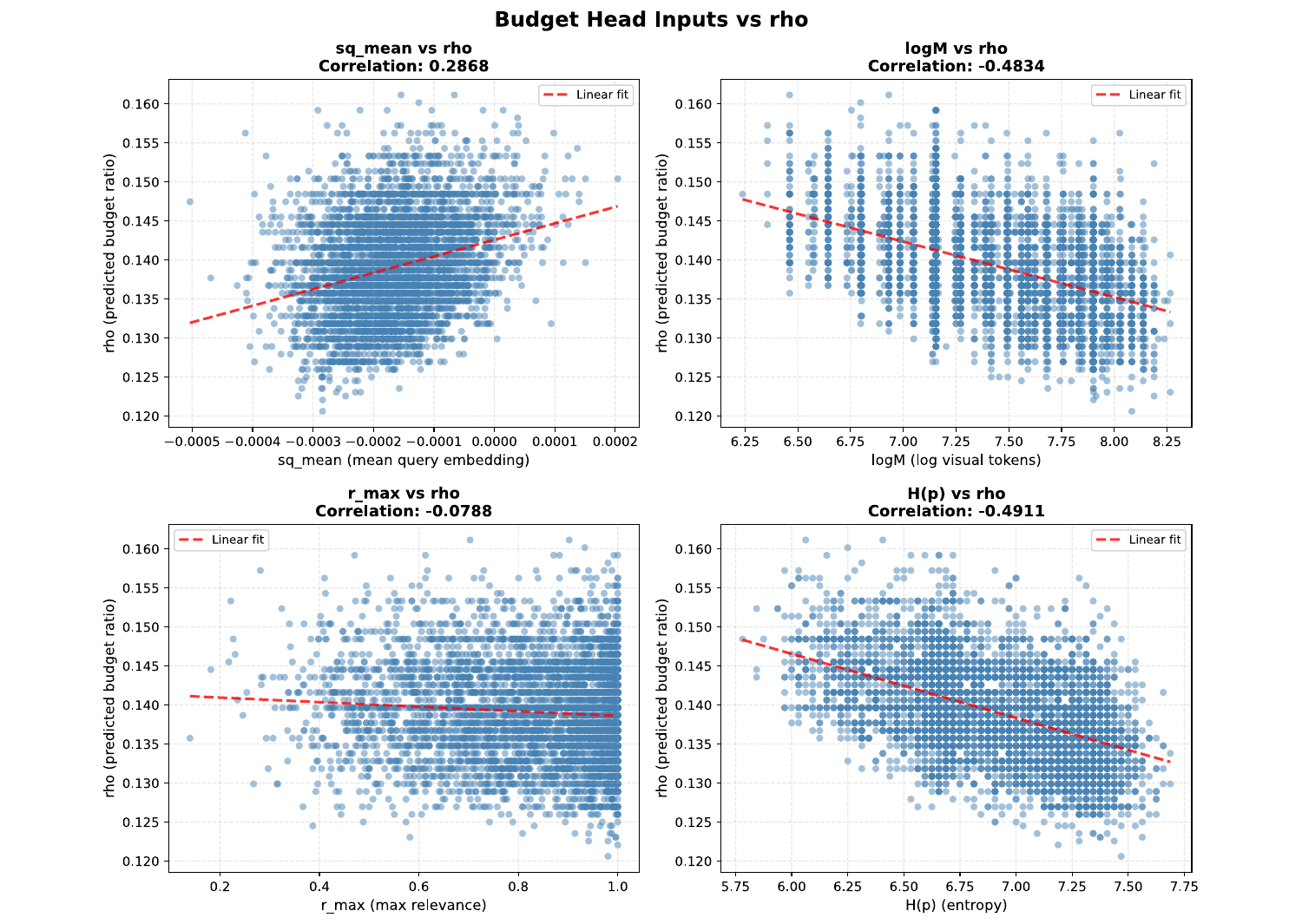}
  \caption{\textbf{Correlation between budget-head inputs and predicted budget ratio $\rho$.} Each subplot shows the relationship between one scalar input to the budget head and the predicted retention ratio $\rho$: (\textbf{top left}) mean query embedding $s_q$, (\textbf{top right}) $\log M$ (logarithm of the number of visual tokens), (\textbf{bottom left}) $r_{\max}$ (maximum query--vision relevance), and (\textbf{bottom right}) $H(p)$ (entropy of the normalised relevance distribution). A red dashed line indicates a linear fit; the reported value in the title is the corresponding correlation coefficient.}
  \label{fig:budget-head-corr}
\end{figure*}

\section{Notation}
We denote an input video by $\mathcal{V} = \{f_t\}_{t=1}^{T}$, where $f_t \in \mathbb{R}^{H \times W \times 3}$ is the $t$-th RGB frame, $T$ is the number of frames, and $H,W$ are the frame height and width. The text query is a token sequence $\mathcal{Q} = (q_1,\dots,q_L)$ of length $L$ drawn from the vocabulary $\mathcal{V}_{\text{text}}$, and the answer sequence is $y$ with length $|y|$; when available, we additionally use a summary token sequence $\Delta = (\delta_1,\dots,\delta_{\alpha})$. After the ViT encoder with patch size $p$ and frame sampling interval $\Delta t$, the video is mapped to a matrix of visual tokens $\mathcal{X} \in \mathbb{R}^{M \times d}$ with $M$ tokens and token dimension $d$, whose $i$-th row is the visual token $x_i$. Our query-aware tokenizer prunes $\mathcal{X}$ to $\mathcal{X}' = [z_1,\dots,z_n]^\top \in \mathbb{R}^{n \times d}$, where $n = \min(\lceil \rho M \rceil, n_{\max})$ is the retained visual-token budget, $n_{\max}$ is a hard cap, and $\rho \in [\rho_{\min}, \rho_{\max}]$ is a predicted retention fraction produced by a budget head $B_{\psi}$.

Cross-attention between text and visual tokens uses $h$ heads with per-head dimension $d_h = d/h$, attention weights $\alpha \in \mathbb{R}^{h \times L \times M}$, relevance scores $r = [r_1,\dots,r_M]$, normalized scores $p_i$, and their entropy $H(p)$. We also define the mean query embedding $s_q$ and the peak relevance $r_{\max} = \max_i r_i$, and use a small constant $\varepsilon$ for numerical stability. The gating module applies a logistic sigmoid $\sigma(\cdot)$ and uses a temperature $\tau_s$ during training; it outputs a one-hot keep/drop vector $y$ via Gumbel–Softmax, a binary keep mask $s_{\text{keep}}$, and the indices $I$ of the top-$n$ kept tokens at inference. The threshold $t$ for gating is chosen by a \textsc{FindThreshold} routine, implemented via Newton’s method with $J$ iterations. We denote by $\mathrm{MHA}_h(\cdot)$ and $\mathrm{FFN}_d(\cdot)$ a lightweight multi-head self-attention block with $h$ heads and a position-wise feed-forward block with hidden size $d$, respectively.

Training is performed in a teacher–student setting, where the teacher $T$ (Qwen2.5-VL) supervises the student $S$ (QTSplus / QTSplus-FT). The tokenizer $g(\cdot)$ maps vision features to a compressed token stream for the language model, and $a^*$ denotes the teacher-validated correct option for video multiple-choice questions. The overall training objective $\mathcal{L}_{\text{total}}$ combines a multiple-choice loss $\mathcal{L}_{\text{mcqa}}$, a sequence-level VQA loss $\mathcal{L}_{\text{vqa}}$, and an instruction-tuning loss $\mathcal{L}_{\text{SFT}}$, together with compute-aware penalties weighted by $(\lambda_t,\lambda_m,\lambda_s)$ and a quadratic regularizer $C_{\rho}$ on the retention fraction. Optionally, a dataset-level target budget $\bar{n}$ is enforced via a dual variable $\alpha$.

Table \ref{tab:notation} shows the details of all the symbols used in this paper. 
\begin{table*}[htb]
\small
\centering
\caption{Notation}
\label{tab:notation}
\setlength{\tabcolsep}{5pt}
\begin{tabular}{@{}p{0.1\linewidth}p{0.35\linewidth}p{0.1\linewidth}p{0.35\linewidth}@{}}
\hline
\textbf{Symbol} & \textbf{Textual explanation} & \textbf{Symbol} & \textbf{Textual explanation} \\
\hline
$\mathcal{V}$ & Video as a sequence of RGB frames $\{f_t\}_{t=1}^T$ (Eq.~(1)) & $f_t$ & $t$-th RGB frame in $\mathbb{R}^{H \times W \times 3}$ \\
$T$ & Number of frames in $\mathcal{V}$ & $H,\,W$ & Frame height and width (pixels) \\
$\mathcal{Q}$ & Text query token sequence $(q_1,\dots,q_L)$ & $q_\ell$ & Embedding of the $\ell$-th query token \\
$L$ & Number of query tokens & $\mathcal{V}_{\text{text}}$ & Text vocabulary \\
$y$ & Answer token sequence & $|y|$ & Length of the answer sequence \\
$\Delta$ & Summary token sequence $(\delta_1,\dots,\delta_\alpha)$ & $\delta_i$ & $i$-th summary token / event \\
$\alpha$ (summary) & Number of summary tokens in $\Delta$ &  &  \\
\hline
$\mathcal{X}$ & Matrix of visual tokens after the ViT (with absolute pos.~enc.), $X\in\mathbb{R}^{M\times d}$ & $x_i$ & $i$-th visual token (row of $\mathcal{X}$) \\
$M$ & Number of visual tokens; $M=\tfrac{T}{\Delta t}\cdot\tfrac{HW}{P^2}$ & $d$ & Token / channel dimension \\
$\Delta t$ & Frame sampling interval (frames) & $P$ & ViT patch size (pixels per side) \\
$\mathcal{X}'$ & Pruned token matrix after selection, $\mathcal{X}'$=$[z_1,\dots,z_n]^\top\in\mathbb{R}^{n\times d}$ & $z_i$ & $i$-th kept token after gating \\
$n$ & Retained visual‑token budget; $n$=$\min(\lceil \rho M\rceil,\,n_{\max})$ & $n_{\max}$ & Hard cap on retained tokens \\
\hline
$h$ & Number of attention heads & $d_h$ & Per‑head dimension ($d/h$) \\
$\alpha$ (attention) & Cross‑attention weights, tensor $\in \mathbb{R}^{h\times L\times M}$ & $r_i$ & Per‑token visual relevance score; $r\in[0,1]^M$ \\
$r$ & Vector of relevance scores $[r_1,\dots,r_M]$ & $p_i$ & Normalized relevance: $p_i=\frac{r_i}{\sum_j r_j + \varepsilon}$ \\
$H(p)$ & Entropy of relevance distribution: $H(p)$=$-\sum_i p_i \log p_i$ & $s_q$ & Mean query embedding: $s_q=\tfrac{1}{L}\sum_{\ell=1}^L q_\ell$ \\
$r_{\max}$ & Peak relevance $\max_i r_i$ & $\varepsilon$ & Small constant for numerical stability \\
\hline
$B_{\psi}$ & Budget‑prediction head (MLP with params $\psi$) & $\rho$ & Predicted retention fraction $\in[\rho_{\min},\rho_{\max}]$ \\
$\rho_{\min},\,\rho_{\max}$ & Lower/upper bounds for $\rho$ & $t$ & Threshold chosen by \textsc{FindThreshold} for the gate \\
$\tau_s$ & Gate temperature used during training & $\sigma(\cdot)$ & Logistic sigmoid function \\
$I$ & Indices of the top‑$n$ tokens at inference & $Z$ & Gated token sequence (temporal order preserved) \\
$y$ (gate) & One‑hot keep/drop vector per token sampled via Gumbel‑Softmax (training) & $s_{\mathrm{keep}}$ & Binary keep mask derived from $y$ \\
\hline
$\mathrm{MHA}_h(\cdot)$ & Lightweight multi‑head self‑attention block (with $h$ heads) & $\mathrm{FFN}_d(\cdot)$ & Position‑wise feed‑forward block (hidden size $d$) \\
$J$ & Iterations in Newton’s method inside \textsc{FindThreshold} &  &  \\
\hline
$T$ (teacher) & Teacher model (Qwen2.5‑VL) & $S$ (student) & Student model (QTSplus / QTSplus‑FT) \\
$g(\cdot)$ & QTSplus tokenizer mapping vision features to a compact stream for the LM & $a^\star$ & Teacher‑validated correct option (for VSCQ) \\
$\mathcal{L}_{\mathrm{mcq}}$ & Multiple‑choice loss (cross‑entropy) & $\mathcal{L}_{\mathrm{vqa}}$ & Sequence‑level distillation loss for VQA \\
$\mathcal{L}_{\mathrm{SFT}}$ & Supervised fine‑tuning loss term used in Eq.~(5) & $\mathcal{L}_{\mathrm{total}}$ & Overall objective (Eq.~(5)) \\
$\lambda_t,\,\lambda_m,\,\lambda_s$ & Weights for compute‑aware penalties in Eq.~(5) & $\bar{\rho}$ & Optional prior mean of $\rho$ (quadratic regularizer) \\
$\bar{n}$ & Optional dataset‑level budget target & $\alpha$ (dual) & Dual variable for enforcing $\bar{n}$ (not to confuse with other $\alpha$’s) \\
\hline
\end{tabular}
\end{table*}

\section{Correlation Analysis of Budget-Head Inputs}
\label{sec:budget-head-corr}

To better understand how the budget head uses its inputs, we compute pairwise correlations between each scalar input
$\{s_q, \log M, r_{\max}, H(p)\}$ and the predicted retention ratio $\rho$ on a held-out set of query--video pairs.
The results are summarised in Fig.~\ref{fig:budget-head-corr}, where each subplot shows the scatter of $(x,\rho)$ values together with a least-squares linear fit and the corresponding correlation coefficient.

Overall, the learned budget head relies most strongly on global video statistics ($\log M$) and the dispersion of relevance scores ($H(p)$), while query-specific features ($s_q$ and $r_{\max}$) play a secondary role.

\paragraph{Mean query embedding $s_q$}
The top-left panel (\emph{sq\_mean vs.\ $\rho$}) shows a mild positive correlation of $0.2868$. This indicates that the semantic content of the query, captured by its mean embedding, has a noticeable but not dominant effect on the allocated budget: different ``types'' of queries tend to receive slightly different average retention ratios, but there is substantial overlap between them.

\paragraph{Video length $\log M$}
In the top-right panel (\emph{logM vs.\ $\rho$}), we observe a moderate negative correlation of $-0.4834$. As the number of visual tokens $M$ increases, the predicted fraction $\rho$ tends to decrease. Since the absolute budget is given by $n = \rho M$, this pattern suggests that the budget head uses video length as a strong control signal, allocating a smaller \emph{fraction} of tokens for longer videos while still allowing the absolute number of retained tokens to grow with $M$.

\paragraph{Peak relevance $r_{\max}$}
The bottom-left panel (\emph{$r_{\max}$ vs.\ $\rho$}) exhibits only a very weak negative correlation of $-0.0788$. The scatter cloud is almost horizontal, implying that the maximum cross-attention relevance alone has limited influence on the predicted budget. In other words, the budget head does not rely heavily on a single \textit{most relevant} token when deciding how many visual tokens to keep.

\paragraph{Relevance entropy $H(p)$}
The bottom-right panel (\emph{$H(p)$ vs.\ $\rho$}) shows the strongest correlation in magnitude, a moderate negative correlation of $-0.4911$. Higher entropy---i.e., more dispersed relevance over tokens—is associated with a lower retention ratio $\rho$, while sharply peaked relevance distributions tend to receive a larger fraction of tokens.
This demonstrates that the budget head is highly sensitive to how concentrated or diffuse the cross-attention scores are, and it uses this dispersion signal as a major factor in its budgeting behaviour.

\paragraph{Summary}
Taken together, these results show that the learned budget head primarily conditions on global structure (video length) and the distributional shape of relevance ($H(p)$), with query semantics and peak relevance providing finer adjustments. The correlations are moderate rather than extreme, which is consistent with the fact that the budget head uses all four inputs jointly through a small MLP; the
pairwise trends in Fig.~\ref{fig:budget-head-corr} thus reflect dominant but not exclusive influences on $\rho$.

\begin{figure}[htb]
  \centering
  \includegraphics[width=\linewidth]{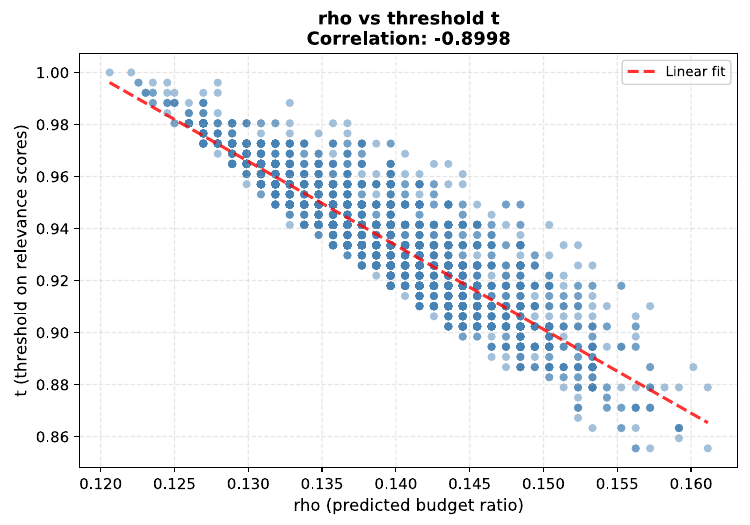}
  \caption{\textbf{Correlation between the predicted budget ratio $\rho$ and the learned threshold $t$ on relevance scores.} Each point corresponds to a query--video pair; the dashed red line shows a least-squares linear fit. We observe a strong negative correlation ($r=-0.8998$): higher budgets $\rho$ correspond to lower thresholds $t$, i.e., the gate becomes more
  permissive and retains tokens with smaller relevance scores.}
  \label{fig:rho-vs-threshold}
\end{figure}
\section{Correlation Analysis between predicted budget $\rho$ and gating threshold $t$}

During training, the soft gate in QTSplus enforces the target budget $\rho$ by choosing a threshold $t$ on the relevance scores $r_i$ such that the expected number of retained tokens matches $\rho M$ (see Algorithm~\ref{alg:findt} in the main
text). For each instance we therefore obtain a scalar pair
$(\rho, t)$, where $\rho$ is predicted by the budget head and $t$ is found by Newton iterations on the sigmoid gate.

Figure~\ref{fig:rho-vs-threshold} confirms that the learned controller behaves in a globally consistent way: as the budget ratio $\rho$ increases from roughly $0.12$ to $0.16$, the threshold $t$ decreases from about $1.0$ to $0.86$. This near-linear, strongly negative correlation indicates that most of the variation in the number of selected tokens is realised by a dynamic shift of the global threshold. The remaining spread around the regression line reflects differences in the shape of the
relevance distribution across videos (e.g., more or less peaked scores), but the overall trend shows that the budget head and the threshold solver are well aligned: larger budgets reliably induce more permissive gates, while smaller budgets tighten the threshold and aggressively prune low-relevance tokens.

\begin{table}[hbt]
\centering
\caption{Training hyperparameters for QTSplus-3B-FT and QTSplus-3B.}
\begin{tabular}{lcc}
\hline
\textbf{Hyperparameter} & \textbf{QTSplus-3B-FT} & \textbf{QTSplus-3B} \\
\hline
model\_max\_length        & 384   & 288   \\
num\_train\_epochs        & 2     & 2     \\
batch\_size               & 1 & 1   \\
gradient\_accumulation    & 1 & 8   \\
learning\_rate            & 1e-6  & 1e-6  \\
weight\_decay             & 0.    & 0.    \\
warmup\_ratio             & 0.01  & 0.01  \\
lr\_scheduler\_type       & cosine & cosine \\
max\_grad\_norm           & 1     & 1     \\
freeze\_vision\_model     & True  & True  \\
freeze\_language\_model   & False & True  \\
$\tau_s$         & 0.5   & 0.5   \\
nmax           & 512   & 256   \\
$\rho_{min}$       & 0.05  & 0.05  \\
$\rho_{max}$       & 0.5   & 0.5   \\
dropout                  & 0.0   & 0.0   \\
reencode       & True  & True  \\
scoring\_layers & 1    & 1     \\
reencode\_layers & 2   & 2     \\
$\lambda_t$                 & 0.2   & 0.1   \\
$\lambda_m$                 & 0.3   & 0.17  \\
$\lambda_s$                 & 0.05  & 0.05  \\
\hline
\end{tabular}
\label{tab:qtsplus-hparams}
\end{table}

\section{Training hyperparameters}
We train \textbf{QTSplus-3B-FT} and \textbf{QTSplus-3B} with a maximum model context length of 384 and 288 tokens, respectively, for 2 epochs with batch size 1. The learning rate is fixed to $1\times 10^{-6}$ with zero weight decay, a cosine learning-rate scheduler, and a warm-up ratio of 0.01. We apply gradient clipping with a maximum gradient norm of 1. The visual encoder is frozen in all experiments, while the language model is fine-tuned for QTSplus-3B-FT and kept frozen for QTSplus-3B. 

The QTSplus module uses a Gumbel straight-through temperature of $\tau_s = 0.5$, minimum and maximum retention fractions $\rho_{\min} = 0.05$ and $\rho_{\max} = 0.5$, and no block dropout. The re-encoding path is enabled with one scoring layer and two re-encoding layers. For QTSplus-3B-FT we use $n_{\max}=512$ and gradient accumulation of 1 step, whereas for QTSplus-3B we use $n_{\max}=256$ and 8 gradient-accumulation steps. The compute-regularization weights in the loss are $(\lambda_t,\lambda_m,\lambda_s)=(0.2, 0.3, 0.05)$ for QTSplus-3B-FT and $(0.1, 0.17, 0.05)$ for QTSplus-3B.

{
    \small
    \bibliographystyle{ieeenat_fullname}
    \bibliography{main}
}

\end{document}